\documentclass[11pt,a4paper]{article}
\usepackage[pdftex]{graphicx}
\usepackage{amsmath}    
\usepackage{amsfonts} 
\usepackage{amssymb}
\usepackage[square,comma,numbers]{natbib}
\usepackage{authblk}

\newenvironment{sciabstract}{%
\begin{quote} \bf}
{\end{quote}}

\begin{document}

\title{Feature-based time-series analysis}


\author[1,2,$\star$]{Ben D. Fulcher}
\affil[1]{Monash Institute for Cognitive and Clinical Neurosciences, Monash University, Melbourne, Victoria, Australia.}
\affil[2]{School of Physics, Sydney University, NSW, Australia.}
\affil[$\star$]{E-mail: ben.d.fulcher@gmail.com}

\date{\today}
\maketitle

\begin{sciabstract}
This work presents an introduction to feature-based time-series analysis.
The time series as a data type is first described, along with an overview of the interdisciplinary time-series analysis literature.
I then summarize the range of feature-based representations for time series that have been developed to aid interpretable insights into time-series structure.
Particular emphasis is given to emerging research that facilitates wide comparison of feature-based representations that allow us to understand the properties of a time-series dataset that make it suited to a particular feature-based representation or analysis algorithm.
The future of time-series analysis is likely to embrace approaches that exploit machine learning methods to partially automate human learning to aid understanding of the complex dynamical patterns in the time series we measure from the world.
\end{sciabstract}

\section{Introduction}
\label{sec12:intro}

\subsection{The time series data type}

The passing of time is a fundamental component of the human experience and the dynamics of real-world processes is a key driver of human curiosity.
On observing a leaf in the wind, we might contemplate the burstiness of the wind speed, whether the wind direction now is related to what it was a few seconds ago, or whether the dynamics might be the similar if observed tomorrow.
This line of questioning about dynamics has been followed to understand a wide range of real-world processes, including in:
seismology (e.g., recordings of earthquake tremors),
biochemistry (e.g., cell potential fluctuations),
biomedicine (e.g., recordings of heart rate dynamics),
ecology (e.g., animal population levels over time),
astrophysics (e.g., radiation dynamics),
meteorology (e.g., air pressure recordings),
economics (e.g., inflation rates variations),
human machine interfaces (e.g., gesture recognition from accelerometer data), and
industry (e.g., quality control sensor measurements on a production line).
In each case, the dynamics can be captured as a set of repeated measurements of the system over time, or a \emph{time series}.
Time series are a fundamental data type for understanding dynamics in real-world systems.
Note that throughout this work we use the convention of hyphenating `time-series' when used as an adjective, but not when used as a noun (as `time series').

\begin{figure}[h]
  \centering
    \includegraphics[width=.9\textwidth]{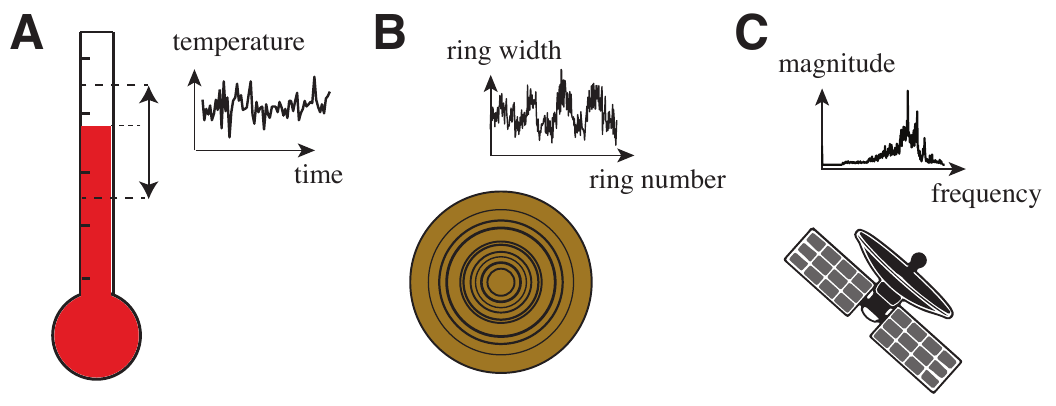}
  \caption[Sequential data]{
  Sequential data can be ordered in many ways, including
  \textbf{A} temperature measured over time (a \emph{time series}),
  \textbf{B} a sequence of ring widths, ordered across the cross section of a tree trunk, and
  \textbf{C} a frequency spectrum of astrophysical data (ordered by frequency).
  All of these sequential measurements can be analyzed by methods that take their sequential ordering into account, including time-series analysis methods.
  }
  \label{fig12:sequential_data}
\end{figure}

In general, time series can be sampled non-uniformly through time, and can therefore be represented as a vector of time stamps, $t_i$, and associated measurements, $x_i$.
However, time series are frequently sampled uniformly through time (i.e., at a constant sampling period, $\Delta t$), facilitating a more compact representation as an ordered vector $x = (x_1,x_2,...,x_N)$, where $N$ measurements have been taken at times $t = (0,\Delta t, 2\Delta t, ..., (N-1)\Delta t)$.
Representing a uniformly-sampled time series as an ordered vector allows other types of real-valued sequential data to be represented in the same way, such as
spectra (where measurements are ordered by frequency),
word length sequences of sentences in books (where measurements are ordered through the text),
widths of rings in tree trunks (ordered across the radius of the trunk cross section),
and even the shape of objects (where the distance from a central point in a shape can be measured and ordered by the angle of rotation of the shape) \cite{keogh2006lb_keogh}.
Some examples are shown in Fig.~\ref{fig12:sequential_data}.
Given this common representation for sequential data, methods developed for analyzing time series (which order measurements by time), can also be applied to understand patterns in any sequential data.
\index{sequential data}

While time series described above are the result of a single measurement taken repeatedly through time, or \emph{univariate} time series, measurements are frequently made from multiple parts of a system simultaneously, yielding \emph{multivariate} time series.
\index{multivariate time series}
Examples of multivariate time series include measurements of the activity dynamics of multiple brain regions through time, or measuring the air temperature, air pressure, and humidity levels together through time.
Techniques have been developed to model and understand multivariate time series, and infer models of statistical associations between different parts of a system that may explain its multivariate dynamics.
Methods for characterizing inter-relationships between time series are vast, including the simple measures of statistical dependencies, like the linear cross correlation, mutual information, and to infer causal (directed) relationships using methods like transfer entropy and Granger causality \cite{Zaremba2014}.
A range of information-theoretic methods for characterizing time series, particularly the dynamics of information transfer between time series, are described and implemented in the excellent Java Information Dynamics Toolkit (JIDT) \cite{Lizier:2014hl}.
Feature-based representations of multivariate systems can include both features of individual time series, and features of inter-relationships between (e.g., pairs of) time series.
However, in this chapter we focus on individual univariate time series sampled uniformly through time (that can be represented as ordered vectors, $x_i$).

\subsection{Time-series characterization}

\begin{figure}[t]
  \centering \includegraphics[width=\textwidth]{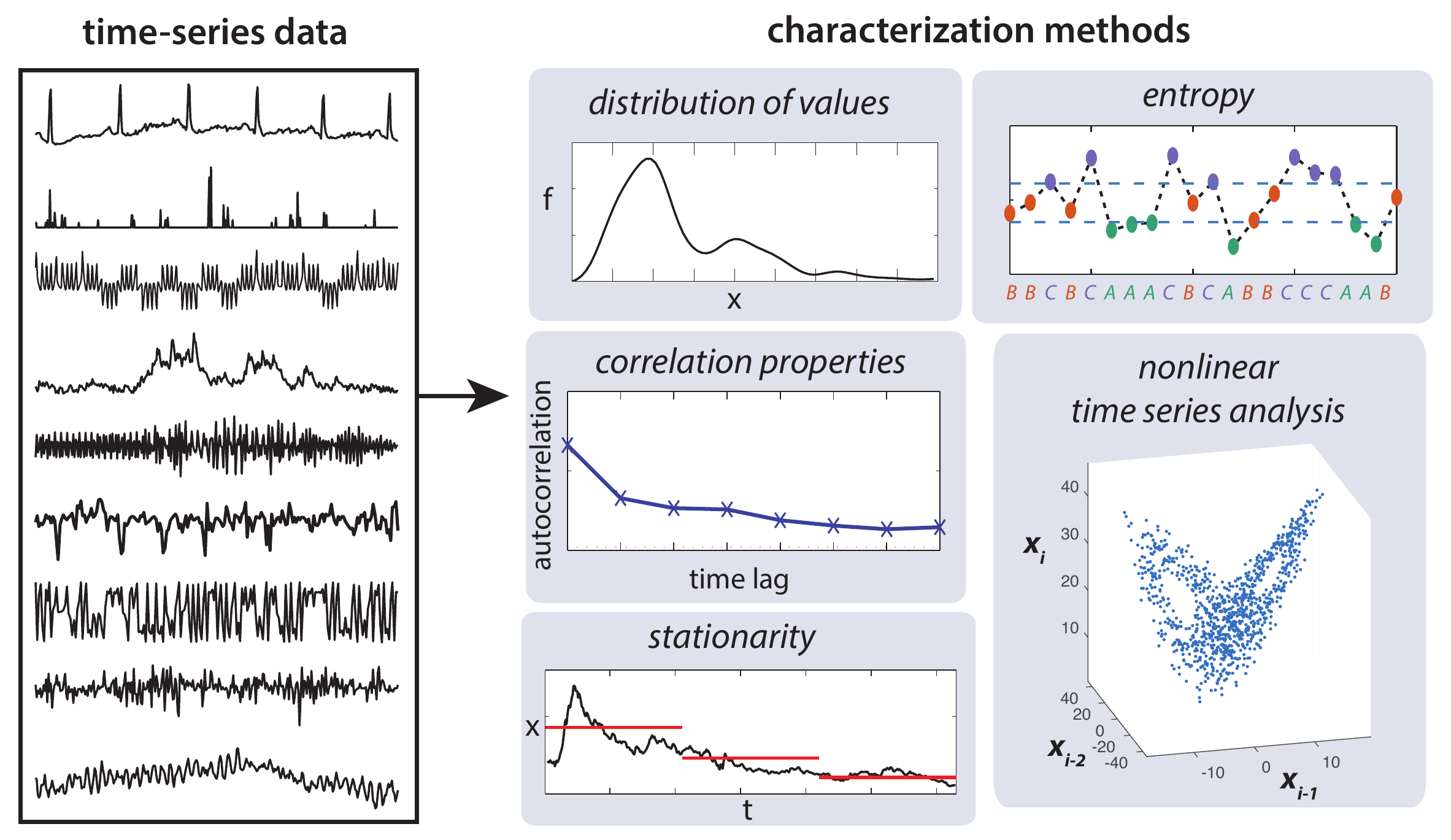}
  \caption[Time-series characterization]{
  \textbf{Time-series characterization.}
  \emph{Left}: A sample of nine real-world time series reveals a diverse range of temporal patterns \cite{Fulcher:2013ft, website-compengine}.
  \emph{Right}: Examples of different classes of methods for quantifying the different types of structure, such as those seen in time series on the left:
  (i) \emph{distribution} (the distribution of values in the time series, regardless of their sequential ordering);
  (ii) \emph{autocorrelation properties} (how values of a time series are correlated to themselves through time);
  (iii) \emph{stationarity} (how statistical properties change across a recording);
  (iv) \emph{entropy} (measures of complexity or predictability of the time series quantified using information theory); and
  (v) \emph{nonlinear time-series analysis} (methods that quantify nonlinear properties of the dynamics).
  }
  \label{fig12:characterization}
\end{figure}

As depicted in the left box of Fig.~\ref{fig12:characterization}, real-world and model-generated time series are highly diverse, ranging from the dynamics of sets of ordinary differential equations simulated numerically, to fast (nanosecond timescale) dynamics of plasmas, the bursty patterns of daily rainfall, or the complex fluctuations of global financial markets.
How can we capture the different types of patterns in these data to understand the dynamical processes underlying them?
Being such a ubiquitous data type, part of the excitement of time-series analysis is the large interdisciplinary toolkit of analysis methods and quantitative models that have been developed to quantify interesting structures in time series, or \emph{time-series characterization}.

We distinguish the characterization of \emph{unordered} sets of data, which is restricted to the distribution of values, and allows questions to be asked like:
`Does the sample have a high mean or spread of values?';
`Does the sample contain outliers?';
`Are the data approximately Gaussian distributed?'.
While these types of questions can also be asked of time series, the most interesting types of questions probe the temporal dependencies and hence the dynamic processes that might underly the data: e.g.,
`How bursty is the time series?';
`How correlated is the value of the time series to its value one second in the future?';
`Does the time series contain strong periodicities?'
Interpreting the answers to these questions in their domain context provides understanding of the process being measured.



Some key classes of methods developed for characterizing time series are depicted in the right panel of Fig.~\ref{fig12:characterization}, and include autocorrelation, stationarity, entropy, and methods from the physics-based nonlinear time-series analysis literature.
Within each broad methodological class, hundreds of time-series analysis methods have been developed across decades of diverse research \cite{Fulcher:2013ft}.
In their simplest form, these methods can be represented as algorithms that capture time-series properties as real numbers, or \emph{features}.
Many different feature-based representations for time series have been developed and been used in applications ranging from time-series modeling, forecasting, and classification.

\subsection{Applications of time-series analysis}
\begin{figure}[h]
  \centering \includegraphics[width=0.8\textwidth]{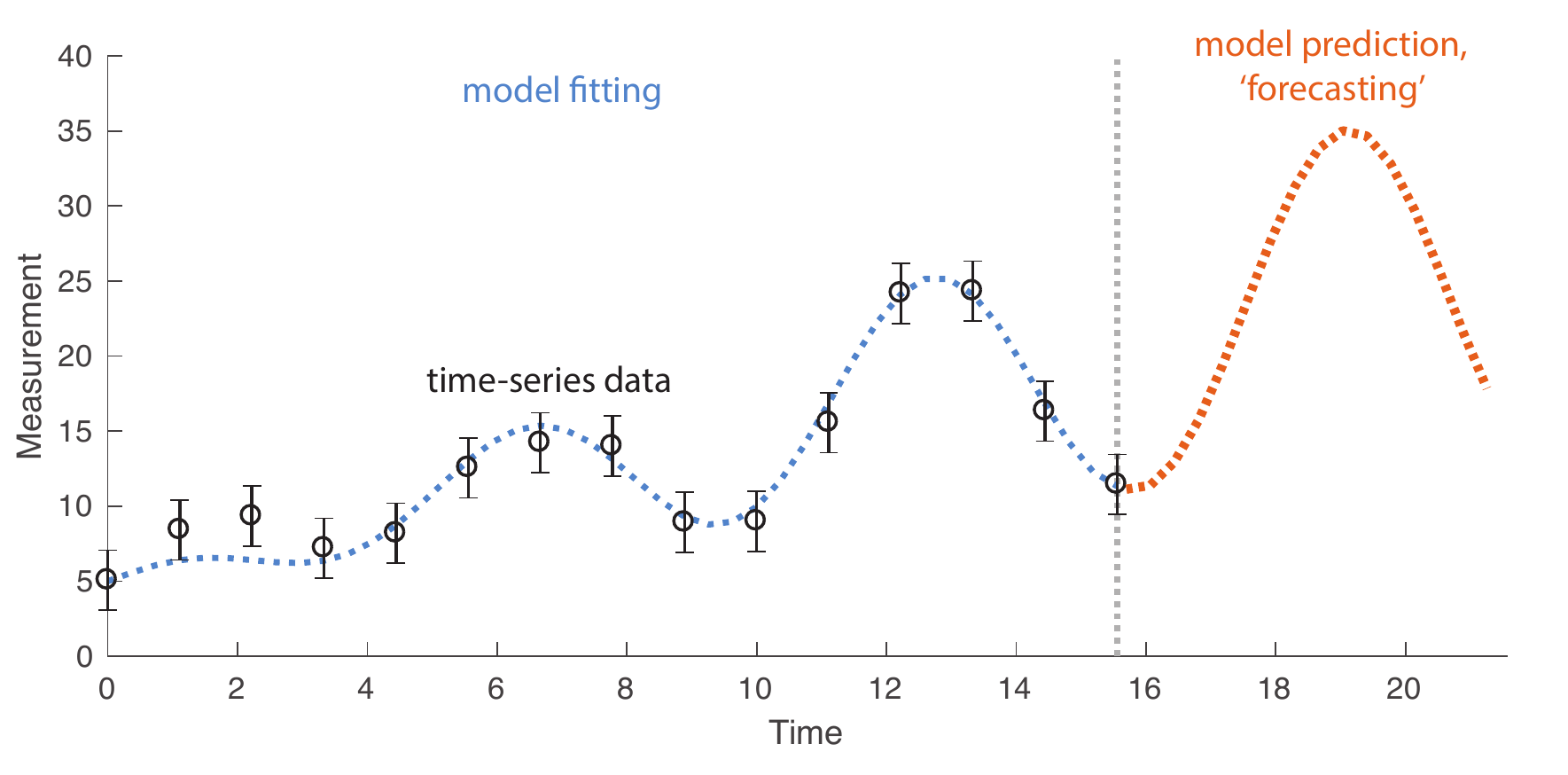}
  \caption[Time-series modeling and forecasting]{
  \textbf{Time-series modeling and forecasting.}
  The figure shows a uniformly sampled time series (black), a model fit (blue), and predictions of the fitted model (orange).
  }
  \label{fig12:forecasting}
\end{figure}

The interdisciplinary reach of the time-series analysis literature reflects the diverse range of problem classes that involve time series.
\emph{Time-series modeling} is perhaps the most iconic problem class.
\index{time-series modeling}
Statistical models can provide understanding of statistical relationships in the data (e.g., autocorrelation structure, seasonality, trends, nonlinearity, etc.),
whereas mechanistic models use domain knowledge to capture underlying processes and interactions as equations that can be simulated in an attempt to reproduce properties of the observed dynamics.
Comparing the quality of fit between different models allows inference of different types of processes that may underly the data.
Models that fit the observed data well can be simulated forward in time to make predictions about the future state of the system, a task known as \emph{forecasting} \cite{hyndman2014}, depicted in Fig.~\ref{fig12:forecasting}.
\index{time-series forecasting}
For example, forecasting could be used to predict the value of a stock in an hour's time, the air temperature at noon tomorrow, or an individual's depression severity in a week.
The range of statistical modeling approaches is vast \cite{chatfield2000, box2015, hyndman2014}, with different applications and data types favoring different approaches, including simple exponential smoothing \cite{GardnerJr2006}, autoregressive integrated moving average (ARIMA) models, generalized auto regressive conditional heteroscedastic (GARCH) models, Gaussian Process models \cite{BrahimBelhouari2004}, and neural networks \cite{Zhang1998}.

Problems in time-series data mining have received much research attention \cite{esling2012time}, and typically center around quantifying the similarity between pairs of time series using an appropriate \emph{representation} and \emph{similarity metric} \cite{Wang12}.
\index{time-series data mining}
This allows one to tackle problems including:
\emph{query by content}, in which known patterns of interest are located in a time series database \cite{faloutsos1994fast};
\emph{anomaly detection}, in which unusual patterns in a time series database are detected, such as unusual (possibly fraudulent) patterns of credit card transactions \cite{weiss2004mining};
\emph{motif discovery}, in which commonly recurring subsequences in a time series are identified \cite{patel2002mining, chiu2003probabilistic};
\emph{clustering}, in which time series are organized into groups of similar time series \cite{keogh2003clustering, Liao05};
and \emph{classification}, in which different labeled classes of time series are distinguished from each other \cite{bakshi1994representation}.

\begin{figure}[h]
  \centering \includegraphics[width=0.8\textwidth]{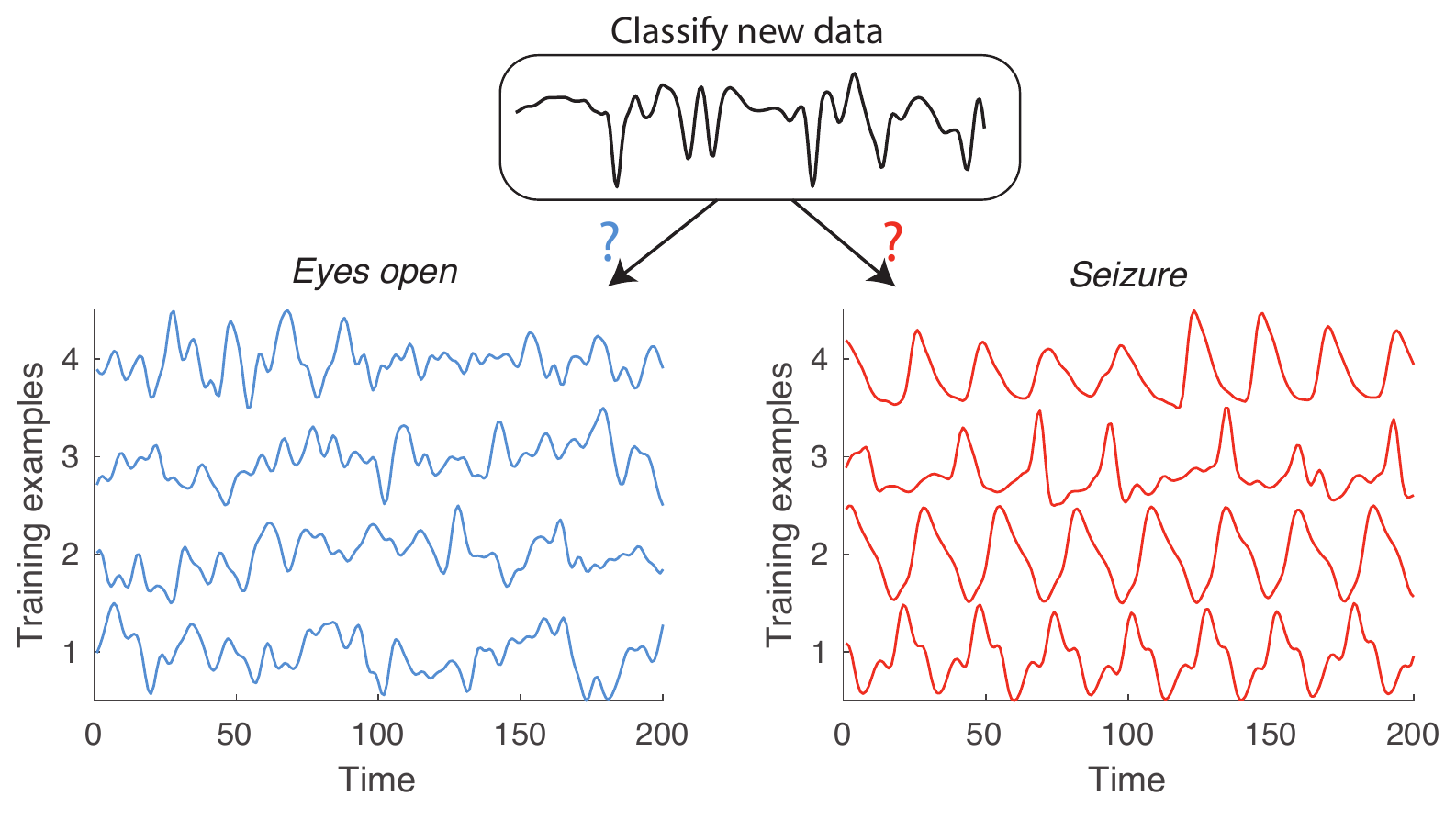}
  \caption[Time-series classification]{
  \textbf{In time-series classification, time series are assigned to categories.}
  In this example, involving EEG signals of two types, the aim is to distinguish data corresponding to `eyes open' (blue) from `seizure' (red) \cite{Andrzejak01}.
    Time-series classification algorithms quantify discriminative patterns in training examples (which the reader may be able to see visually), and use them to classify new data into one of the predefined groups.
  }
  \label{fig12:classification}
\end{figure}

Time-series classification, depicted in Fig.~\ref{fig12:classification}, is a much-studied problem that we revisit throughout this chapter.
\index{time-series classification}
The figure depicts the goal of classifying a new time series as being measured from a subject in one of two states: (1) at rest with `eyes open', or (2) during a `seizure' \cite{Andrzejak01}.
The aim is to learn the most discriminative differences between the classes from labeled training examples, and use this information to accurately classify new data.
The same framework can be used to detect faults on a production line using time-series sensor recordings (`safe' versus `fault'), or diagnose heart rate dynamics of a patient with congestive heart failure from a healthy control (`healthy' versus `heart failure').
Note that while most applications have considered a categorical target variable, this problem can also be placed in a \emph{regression} framework when the target variable is continuous \cite{Fulcher:2013ft, Christ2017} (see Sec.~4 of the supplementary text of \cite{Fulcher:2013ft}).
For example, rather than predicting whether a patient has `high' or `low' blood pressure (as in a classification framework), this can allow the direct prediction of blood pressure from a physiological time series.

\section{Feature-based representations of time series}
\label{sec12:timeSeriesFeatures}
In this section, we motivate feature-based representations of time series using the problem of defining a measure of similarity between pairs of time series, which is required for many applications of time-series analysis, including many problems in time-series data mining \cite{Keogh03, Liao05, Wang12}.
As we saw above in time-series classification, new time series are classified by matching them to the most similar training time series and then inferring their class label (cf. Fig.~\ref{fig12:classification}).
What is meant by `similar' in the context of this problem---and hence the motivation for defining a similarity metric---determines the classification results.
Despite the pursuit of algorithms that perform `best' (on general datasets), the No-Free-Lunch theorem identifies that good performance of an algorithm on any class of problems is offset by poorer performance on another class \cite{wolpert1996, Wolpert97}.
\index{No-Free-Lunch theorem}
In the context of time-series representations, this implies that there is no `best' representation or similarity metric in general; the most useful measure depends on the data and the questions being asked of it \cite{bagnall2017simulated}.
For example, it may sometimes be useful to define similar time series as those with similar frequency content, while for other applications it may be more useful to define and quantify a burst rate, and define similar time series as those with similar burst rates.

\begin{figure}[h]
  \centering \includegraphics[width=\textwidth]{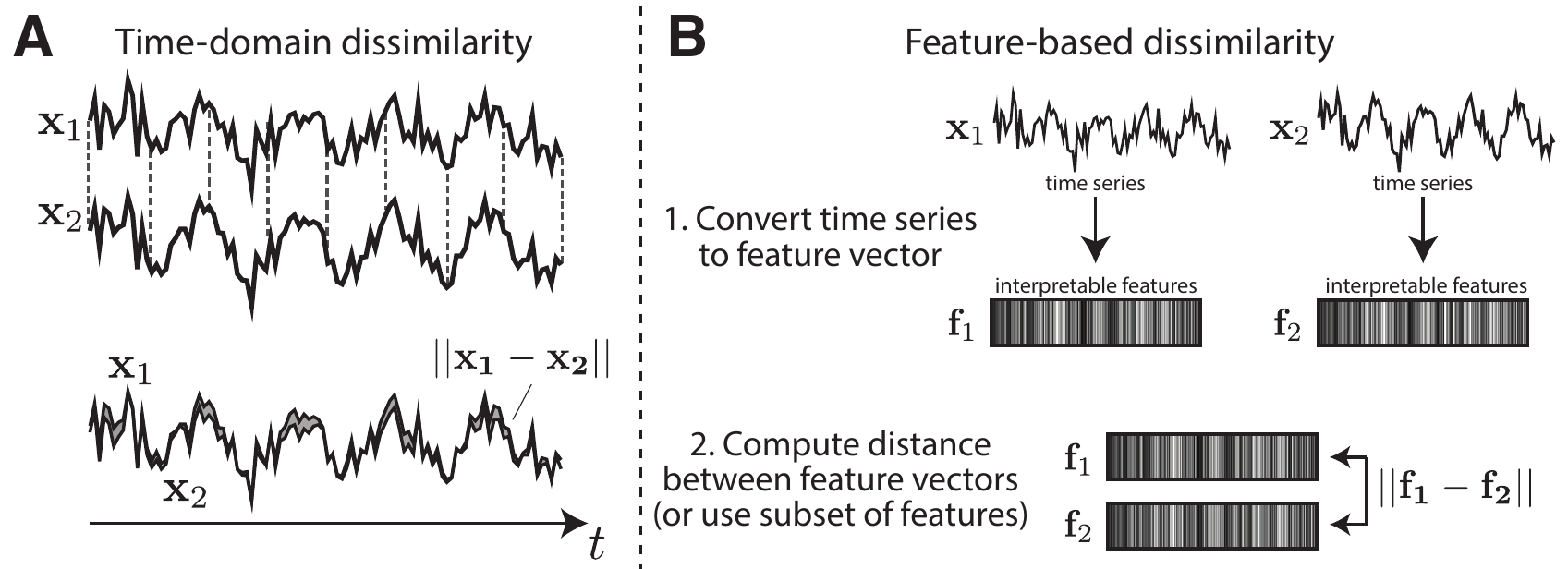}
  \caption[Similarity measures for pairs of time series]{
  \textbf{Time-domain and feature-based dissimilarity measures for pairs of time series.}
  \textbf{A} Dissimilarity between two time series, $\mathbf{x}_1$ and $\mathbf{x}_2$, can be computed in the time domain as a simple Euclidean distance, e.g., $||\mathbf{x}_1-\mathbf{x}_2||_2$.
This judges two time series as similar that have similar values at similar times.
  \textbf{B} An alternative similarity metric involves computing a set of interpretable features from each time series (such as properties of their distribution, correlation properties, etc.), and then computing the dissimilarity between their features, as $||\mathbf{f}_1-\mathbf{f}_2||_2$.
This judges two time series as similar that have similar sets of global properties.
  }
  \label{fig12:dissimilarityMeasures}
\end{figure}

For time series of equal length, perhaps the simplest measure of time-series similarity captures how similar time-series values are across the time period, depicted in Fig.~\ref{fig12:dissimilarityMeasures}A.
This approach judges two time series as similar that have similar values at similar times, and can be quantified using simple metrics like the Euclidean distance between the two time series vectors.
Many more sophisticated similarity measures that capture the similarity of time-series values through time have been developed, including `elastic' distance measures like dynamic time warping (DTW) that do not require time series to be aligned precisely in time \cite{Berndt94, Wang12, Ratanamahatana04, batista2014cid, lines2015time}.
\index{dynamic time warping (DTW)}
Across typically studied time-series classification problems, DTW combined with one-nearest-neighbor classification (i.e., classifying new time series by the time series in the training set with the smallest DTW distance) can yield high classification performance \cite{bagnallreview}.

While optimizing performance (e.g., classification accuracy) is the main objective in many real-world applications of machine learning, providing understanding of why performance is good is often preferable, especially for applications in scientific research.
Although comparing sequential values through time can yield high classification accuracy for some applications, nearest neighbor classifiers involve computing distances from a new time series to many training time series (which can be computationally expensive) and, most importantly, does not provide deeper understanding of the characteristics of time series in different classes that drive successful classification.
An alternative is to use a \emph{feature-based representation}, which reframes the problem in terms of interpretable features.
Timmer et al.~wrote: \emph{``The crucial problem is not the classificator function (linear or nonlinear), but the selection of well-discriminating features. In addition, the features should contribute to an understanding [...]''} \cite{Timmer93}.
Similar sentiments have been mirrored by others, who have argued that research aiming to improve classification performance should focus on transformations of time series into useful (e.g., feature-based) spaces, rather than trying to develop and apply new and complex classifiers in spaces that may not best represent the data for the desired application \cite{bagnall2012transformation, harvey2015automated, Fulcher:2014uo, Bagnall:2015}.
A feature-based approach to time-series classification is illustrated in Fig.~\ref{fig12:dissimilarityMeasures}B, in which each time series is converted to a set of global features (such as measures of its trend, entropy, distribution), which are used to define the similarity between pairs of time series \cite{Fulcher:2014uo}.
Understanding which feature-based representations provide good performance for a given task can provide conceptual understanding of the properties of the data that drive accurate decision making, information that can in turn be used to inform domain understanding and motivate future experiments.

Feature-based representations of time series can be used to tackle a wide range of time-series analysis problems in a way that provides interpretability, with the choice of feature-based representation determining the types of insights that can be gained about the problem at hand.
For example, in the case of global features derived from time-series analysis algorithms, understanding comes from the features which encode deeper theoretical concepts (like entropy, stationarity, or Fourier components, described in Sec.~\ref{sec:global_feature_examples} below) -- e.g., the researcher may learn that patients with congestive heart failure have heart rate intervals with lower entropy.
If features are instead derived from the shapes of time-series subsequences, understanding comes from the time-series patterns in discriminatory time intervals -- e.g., the researcher may learn that patients with congestive heart failure have characteristic shapes of ECG fluctuations following the onset of atrial depolarization.
As different time-series similarity metrics are introduced through the following sections, the reader should keep in mind the benefits of each representation in terms of the understanding it can provide to a given problem.

\section{Global features}

Global features refer to algorithms that quantify patterns in time series across the full time interval of measurement (e.g., rather than capturing shorter subsequences).
Global features can thus distill complicated temporal patterns that play out on different timescales and can be produced by a variety of complex underlying mechanisms, into interpretable low-dimensional summaries that can provide insights into the generative processes underlying the time series.
Representing a time series in terms of its properties judges two time series to be similar that have similar global properties, and can thereby connect dynamics in time series to deeper theoretical ideas.
Global features can be applied to time series of variable lengths straightforwardly, and allows a time-series dataset to be represented as a time series (rows) $\times$ features (columns) data matrix, which can form the basis of traditional statistical learning (e.g., machine learning) methods.
An example is shown in Fig.~\ref{fig12:schematic_hctsa}.

\subsection{Examples of global features}\label{sec:global_feature_examples}

There is a vast literature of time-series analysis methods for characterizing time-series properties \cite{Fulcher:2013ft} that can be leveraged to extract interpretable features from a time series.
In this section we list some examples of specific global time-series features from some of the major methodological classes to give a flavor for the wide, interdisciplinary literature on time-series analysis \cite{Fulcher:2013ft} (cf. Fig.~\ref{fig12:characterization}).

Simple measures of the \emph{distribution} of time-series values (which ignore their time-ordering) can often be informative.
Examples include the mean, variance, fits to distribution types (e.g., Gaussian), distribution entropy, and measures of outliers.
A simple example is the (unbiased) sample variance:
\begin{equation}
    s^2_x = \frac{1}{N-1}\sum_{i=1}^N (x_i - \bar{x})^2,
\end{equation}
for a time series, $x$, of length $N$ with mean $\bar{x}$.
Note how the variance is independent of the ordering of values in $x$.

A \emph{stationary} time series is produced from a system with fixed and constant parameters throughout the recording (or probability distributions over parameters that do not vary across the recording).
Measures of stationarity capture how temporal dependences vary over time.
For example, the simple $\mathrm{StatAv}$ metric provides a measure of mean stationarity \cite{pincus1993}:
\begin{equation}
    \mathrm{StatAv}(\tau) = \frac{\mathrm{std}(\{\overline{x_{1:w}},\overline{x_{w+1:2w}},...,\overline{x_{(m-1)w+1:mw}}\})}{\mathrm{std}(x)},
\end{equation}
where the standard deviation is taken across the set of means computed in $m$ non-overlapping windows of the time series, each of length $w$.
Time series in which the mean in windows of length $w$ vary more than the full time series as a whole have higher values of StatAv at this timescale relative to time series in which the windowed means are less variable.
\index{time-series stationarity}

\emph{Autocorrelation} measures the correlation between time-series values separated by a given time-lag.
The following provides an estimate:
\begin{equation}\label{eq:acf}
    C(\tau) = \langle x_t x_{t+\tau}\rangle = \frac{1}{s^2_x(N-\tau)}\sum_{t=1}^{N-\tau}(x_t - \bar{x})(x_{t+\tau} - \bar{x}),
\end{equation}
at a time lag $\tau$, for a time series, $x$, with variance $s^2_x$ and mean $\bar{x}$.
\index{autocorrelation}
Nonlinear generalizations can also be computed, for example, as the automutual information \cite{Jeong01, kantz2004}.
Apart from autocorrelation values at specific time lags, other features aim to quantify the structure of the autocorrelation function, such as the earliest time lag at which it crosses zero.

The (discrete-time) \emph{Fourier transform} allows a time series to be represented as a linear combination of frequency components, with each component given by:
\begin{equation}
    \tilde{x}_k = \frac{1}{\sqrt{N}} \sum_{n=1}^{N} x_n e^{2\pi ikn/N},
\end{equation}
where the real and complex parts of $\tilde{x}_k$ encode the amplitude and phase of that component, $e^{2\pi ikn/N}$, for frequencies $f_k = k/N\Delta t$, where $\Delta t$ is the sampling interval.
\index{discrete-time Fourier transform}
Other basis function decompositions, such as \emph{wavelet decompositions} use a wavelet basis set under variations in temporal scaling and translation, to capture changes in, e.g., frequency content through time (using a Morlet wavelet) \cite{graps1995introduction}.
\index{wavelet transform}

Given that linear systems of equations can only produce exponentially growing (or decaying) or (damped) oscillatory solutions, irregular behavior in a linear time series must be attributed to a stochastic external drive to the system.
An alternative explanation is that the system displays nonlinearity; deterministic nonlinear equations can produce irregular (chaotic) dynamics which can be quantified using methods from the physics-based \emph{nonlinear time-series analysis} literature \cite{kantz2004}.
These algorithms are typically based on a phase space reconstruction of the time series, e.g., using the method of delays \cite{Takens81}, and include measures of the Lyapunov exponent, correlation dimension, correlation entropy, and others.
\index{nonlinear time-series analysis}

\emph{Entropy} measures are derived from information theory and have been used to quantify predictability in a time series, with specific examples including Approximate Entropy (ApEn) \cite{Pincus91a}, Sample Entropy (SampEn) \cite{Richman00}, and Permutation Entropy (PermEn) \cite{Bandt02}.
\index{time-series entropy}
For example, ApEn$(m,r)$ is defined as the logarithmic likelihood that the sequential patterns of the data (of length $m$) that are close to each other (within a threshold, $r$), will remain close for the next sample, $m+1$.
An unstructured time series has high ApEn, and can be distinguished from a regular deterministic signal, which has a higher probability of similar sequences remaining similar (and thus low ApEn).
Representing time series as a time-delay embedding in terms of a set of vectors $u_m$, which each contain $m$ consecutive values of the time series \cite{kantz2004}, ApEn$(m,r)$ is defined as:
\begin{equation}
    \mathrm{ApEn}(m,r) = \Phi^m(r)-\Phi^{m+1}(r),
\end{equation}
where
\begin{equation}
    \Phi^m(r) = \frac{1}{N-m+1} \sum_{i=1}^{N-m+1} \ln C_r^m(i),
\end{equation}
and
\begin{equation}
    C_i^m(r) = \frac{A_i}{N-m+1}.
\end{equation}
$A_i$ is the number of vectors, $u_m$, within a distance $r$ of $u_m$ (for a given distance function) \cite{Pincus91a}.

\emph{Scaling} algorithms capture the power-law scaling of time-series fluctuations over different timescales, as would be produced by a self-affine or fractal process \cite{kantz2004}.
A stationary time series with long-range correlations can be interpreted as increments of a diffusion-like process and integrated (as a cumulative sum through time) to form a self-similar time series, i.e., a time series that statistically resembles itself through rescaling in time.
\index{time-series scaling}
So-called `short-range correlations', Eq.~\eqref{eq:acf}, decay exponentially as $C(\tau) \sim e^{-\lambda \tau}$; whereas `long-range correlations' decay as a power law, $C(\tau) \sim \tau^{-\gamma}$.
\index{long-range correlations}
Detrended fluctuation analysis \cite{Peng95a, Kantelhardt01} is one method that allows estimation of the correlation exponent, $\gamma$, via the fluctuation function:
\begin{equation}
    F(s) = \left[ \frac{1}{N} \sum_{j=1}^N z_i(s)^2 \right]^{1/2},
\end{equation}
which is computed over a range of scales, $s$, for a fluctuation function $z_i(s) = y_i - y_{\mathrm{fit},i}$, for the integrated time series, $y$, non-overlapping subsequences labeled with the index $i$, and (e.g., linear or quadratic) trends, $y_{\mathrm{fit},i}$, subtracted from each subsequence.
\index{time-series scaling!detrended fluctuation analysis (DFA)}
Scaling as $F(s) \sim s^\alpha$ quantifies long-range power law scaling of time-series fluctuations, with $\alpha$ related to the correlation exponent, $\gamma$, as $\alpha = 1-\gamma/2$ \cite{Kantelhardt01}.

Statistical \emph{time-series models} can be fit to data to better understand complex dynamical patterns.
The range of models is extensive, and includes exponential smoothing models, autoregressive models, moving average models, and Gaussian process models \cite{chatfield2000, box2015, hyndman2014}.
For example, an exponential smoothing model makes predictions about future values of a time series using a weighted sum of its past values, using a smoothing parameter, $\alpha$:
\begin{equation}
    \hat{x}_{t+1} = \alpha x_t + \alpha (1-\alpha) x_{t-1} + \alpha (1-\alpha)^2 x_{t-2} + ...,
\end{equation}
for a prediction of the value at the next time-step, $\hat{x}_{t+1}$, with $0 < \alpha \leq 1$ such that values further into the past are weighted (exponentially) less in the prediction \cite{hyndman2014}.
Many different types of features can be extracted from time-series models, including the model parameters (e.g., the optimal $\alpha$ of an exponential smoothing model), and goodness of fit measures (e.g., as the autocorrelation of residuals).
\index{time-series forecasting}
\index{time-series forecasting!exponential smoothing}

\subsection{Massive feature vectors and highly comparative time-series analysis}

Given the large number of global features that can be used to characterize different properties of a time series, the selection of which features best capture the relevant dynamics of a given dataset typically follows the expertise of a data analyst.
Examples of using manually curated feature sets are numerous, and include:
\begin{itemize}
    \item Timmer et al.~\cite{Timmer93} characterized hand tremor time series using a variety of time- and frequency-domain features; 
    \item Nanopoulos et al.~\cite{Nanopoulos01} used the mean, standard deviation, skewness, and kurtosis of the time series and its successive increments as features to classify synthetic control chart patterns used in statistical process control;
    \item M\"orchen~\cite{Morchen03} used features derived from wavelet and Fourier transforms to classify classes within each of 17 time-series datasets, including buoy sensor data, ECGs, currency spot prices, and gene expression;
    \item Wang et al.~\cite{Wang06} used thirteen features containing measures of trend, seasonality, periodicity, serial correlation, skewness, kurtosis, chaos, nonlinearity, and self-similarity to represent time series, an approach that has since been extended to multivariate time series \cite{Wang07};
    \item Bagnall et al.~\cite{bagnall2012transformation} represented time series as a power spectrum, autocorrelation function, and in a principal components space, demonstrating the potential for a large increase in classification accuracy for feature-based representations (and leveraging different representations together in an ensemble).
\end{itemize}

In each application listed above---as is typical of data analysis in general---the choice of which features to use to characterize a time series is subjective and non-systematic.
Thus, it is difficult to determine whether the features selected by one researcher might differ had they been selected by a different researcher, and therefore whether the feature set presented for a given task is better than existing alternatives \cite{Fulcher:2013ft, Fulcher:2014uo, Fulcher:2016ht}.
The problem is well illustrated by the problem of distinguishing EEG time series during a seizure, for which existing studies had used features derived from a discrete wavelet transform \cite{Subasi10} or a neural network classifier using a multistage nonlinear pre-processing filter with Lyapunov exponents, relative spike amplitude and spike occurrence frequency features \cite{Guler05}; although implicit, it is difficult to establish whether these complicated methodological approaches outperform simpler alternatives.
Indeed, it has been shown that a threshold on the simple standard deviation computed for each time series provides comparable classification performance on this problem, undermining the need for computing nonlinear features or using complex classification algorithms \cite{Fulcher:2013ft}.

Comprehensive methodological comparison---which is rarely done, even on a small scale \cite{Keogh03}---is required to determine whether alternative feature-based representation of time series could be simpler and/or outperform a manually-selected representation.
A difficulty in performing such a comparison is the vast, interdisciplinary nature of the time-series analysis literature, that has been developed over many decades and spanning methods and models used to inform policy decisions from economic time series and those developed to diagnose disease from biomedical time series.
Could it be possible to distill decades of time-series analysis research spanning thousands of studies, datasets, and applications into a unified feature set that would allow us to judge progress through comprehensive methodological comparison?
Such a resource would not only allow us to partially automate the comparison of features, but would also allow us to understand previously uncharacterized methodological connections between an interdisciplinary literature and to judge whether newly-developed methods for time-series analysis outperform existing alternatives (and understand what types of time-series analysis problems they perform well on).

\begin{figure}[h]
  \centering
    \includegraphics[width=\textwidth]{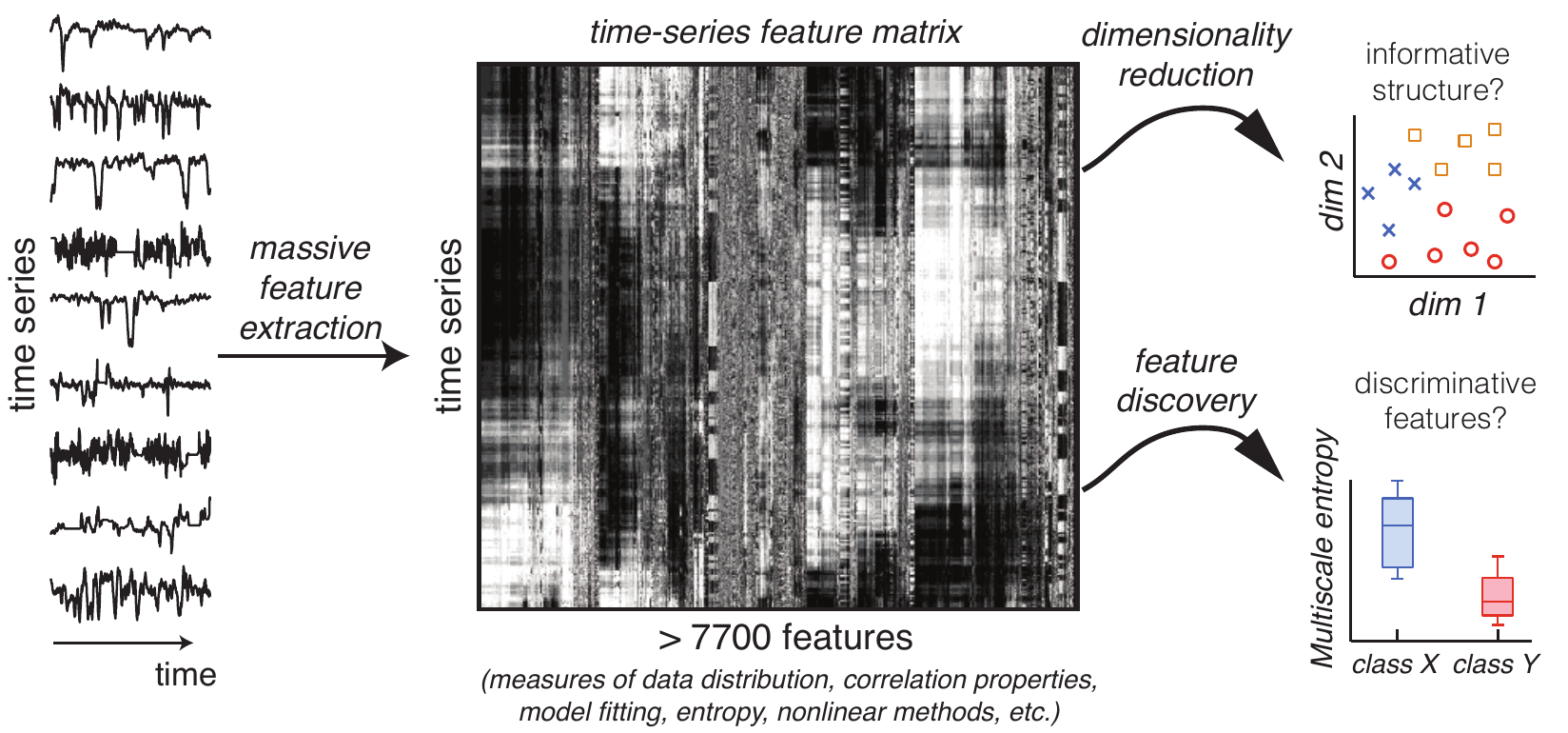}
  \caption[Massive feature extraction for highly comparative time-series analysis]{
  \textbf{Massive feature extraction for time-series datasets.}
  In this example, we show how a time-series dataset (left) can be converted to a time series $\times$ feature matrix.
  Each feature captures a different interpretable property of each time series, and the resulting feature matrix can be used to enable a range of analysis tasks, including visualizing low-dimensional structure in the dataset \cite{Fulcher:2013ft} or learning discriminative features for time-series classification \cite{Fulcher:2014uo}.
  Code for computing features and performing a range of analyses and visualizations is in the \emph{hctsa} software package \cite{Fulcher:2016ht}.
  }
  \label{fig12:schematic_hctsa}
\end{figure}

The problem of unifying and structuring the interdisciplinary literature on global time-series features was addressed in 2013 by Fulcher, Little, and Jones \cite{Fulcher:2013ft}, who collected and implemented hundreds of methods for characterizing time series from across science into a consolidated framework, operationalizing each method as a feature (or set of features), and comparing the behavior of over $>$9\,000 such features using their behavior on a large dataset of $>$30\,000 empirical time series collected from across science.
As the approach involves comprehensive comparison across the time-series analysis literature, it was termed \emph{highly comparative time-series analysis}.
\index{highly comparative time-series analysis}
In this framework, time series (rows of the matrix in Fig.~\ref{fig12:schematic_hctsa}) are represented as diverse feature vectors of their properties measured using thousands of time-series analysis methods, while time-series analysis methods (columns of the matrix in Fig.~\ref{fig12:schematic_hctsa}) are represented in terms of their behavior across a wide range of empirical time series.
This representation of time-series data in terms of their properties, and time-series analysis methods in terms of their empirical behavior, facilitates a range of new approaches to time-series analysis, including:
\begin{itemize}
    \item \emph{Contextualize empirical time-series data}: using a feature-based representation of time series to find clusters of similar time series (e.g., to automatically visualize structure a time-series dataset), or search for similar types of time-series data to a given target (e.g., to find model-generated time series with similar properties to real data).
    \item \emph{Contextualize algorithms for time-series analysis} in terms of their cross-disciplinary interrelationships: using the behavior of time-series analysis methods across a large number of time series to find clusters of similar analysis methods (e.g., to organize an interdisciplinary literature), or search for similar types of methods to a given target method (e.g., to automatically connect features developed in different disciplines).
    \item \emph{Automate the selection of useful feature-based representations of time series}: searching across a comprehensive feature set for supervised learning tasks such as classification or regression.
\end{itemize}


When analyzing a specific dataset, the highly comparative approach allows a feature-based representation for time series to be learned systematically, tailoring it to the problem at hand using the empirical behavior of a large number of analysis methods (shown schematically in Fig.~\ref{fig12:schematic_hctsa}).
This process can be used to guide the data analyst in their task of selecting and interpreting relevant analysis methods.
Applications of the highly comparative approach to supervised and unsupervised time-series analysis problems include:
classifying time series using a large range of time-series data mining datasets \cite{Fulcher:2014uo};
diagnosing phoneme audio recordings from individuals with Parkinson's disease \cite{Fulcher:2013ft};
automatically retrieving and organizing a relevant literature of features for distinguishing heart rate interval sequences of patients with congestive heart failure \cite{Fulcher:2013ft};
labeling the emotional content of speech \cite{Fulcher:2013ft};
distinguishing earthquakes from explosions \cite{Fulcher:2013ft};
projecting a database of EEG recordings into a low-dimensional feature space that revealed differences in seizure-related states \cite{Fulcher:2013ft};
estimating the scaling exponent of self-affine time series, the Lyapunov exponent of Logistic Map time series, and the noise variance added to periodic time series \cite{Fulcher:2013ft};
deciding whether to intervene during labor on the basis of cardiotocogram data \cite{Fulcher:2012gj};
distinguishing \emph{C. elegans} genotypes from movement speed dynamics \cite{Brown:2013ew, Fulcher:2016ht};
learning differences between male and female flies during day or night from tracking their movement \cite{Geissmann:2017kp, Fulcher:2016ht}; and
determining the dynamical correlates of brain connectivity from fMRI data in anesthetized mice \cite{Sethi:2017bx}.
In these diverse disciplinary applications, the highly-comparative approach:
(i) selects features based on their performance on a time-series dataset in a systematic and unbiased way,
(ii) the selected features facilitate interpretable insights into each time-series dataset,
(iii) features are often selected from unexpected literatures (drawing attention to novel features),
(iv) classifiers are often constructed using a novel combination of interdisciplinary features (e.g., combining features developed in economics with others developed in biomedical signal processing),
(v) classifiers have high accuracy, comparable to state-of-the-art approaches,
(vi) the concise, low-dimensional feature-based representations of time series aid data mining applications.

A Matlab-based computational framework for evaluating a refined set of $> 7\,700$ interpretable global time-series features, as well as a suite of computational and analysis functions for applying the results to time-series classification tasks, for example, is available as the software implementation, \emph{hctsa} \cite{Fulcher:2016ht} at \texttt{github.com/benfulcher/hctsa}.
The work is accompanied by an interactive website for comparing data and methods for time-series analysis \cite{website-compengine}.
On a smaller scale to \emph{hctsa}, the related python-based package, \emph{tsfresh}, includes implementations of hundreds of features and includes univariate relevance scoring feature selection methods designed around applications in data mining \cite{Christ:2016}.
\index{highly comparative time-series analysis}
\index{highly comparative time-series analysis!\emph{hctsa} software}

Comparative approaches to selecting global features for time series, described above, are limited to features that have already been developed and devised, i.e., there is no scope to devise completely new types of features for a given dataset.
Automated feature construction for time series, such as the genetic programming (GP)-based approach \emph{Autofead} (using combinations of interpretable transformations, like Fourier transforms, filtering, nonlinear transformations, and windowing) are powerful in their ability to adapt to particular data contexts to generate informative features \cite{harvey2015automated}.
However, features generated automatically in this way can be much more difficult to interpret, as they do not connect the data to interpretable areas of the time-series analysis literature.



\section{Subsequence features}
The previous section outlined how time series can be converted from a sequential set of measurements to a feature vector that captures interpretable global dynamical properties of a time series.
However, for some classification problems, time-series properties may differ only within a specific time interval such that a most efficient representation captures these more temporally specific patterns.
We refer to a subsequence, $s$, of length $l$, taken from a time series, $x$, of length $N$, as $s = (x_k,x_{k+1},...,x_{k+l-1})$, for $1 \leq k \leq m-l+1$, where $l \leq N$.
Different approaches to quantifying subsequences and extracting meaningful and interpretable features from them depend on the application and are summarized through this section.

\subsection{Interval features}
\begin{figure}[h]
  \centering
    \includegraphics[width=.7\textwidth]{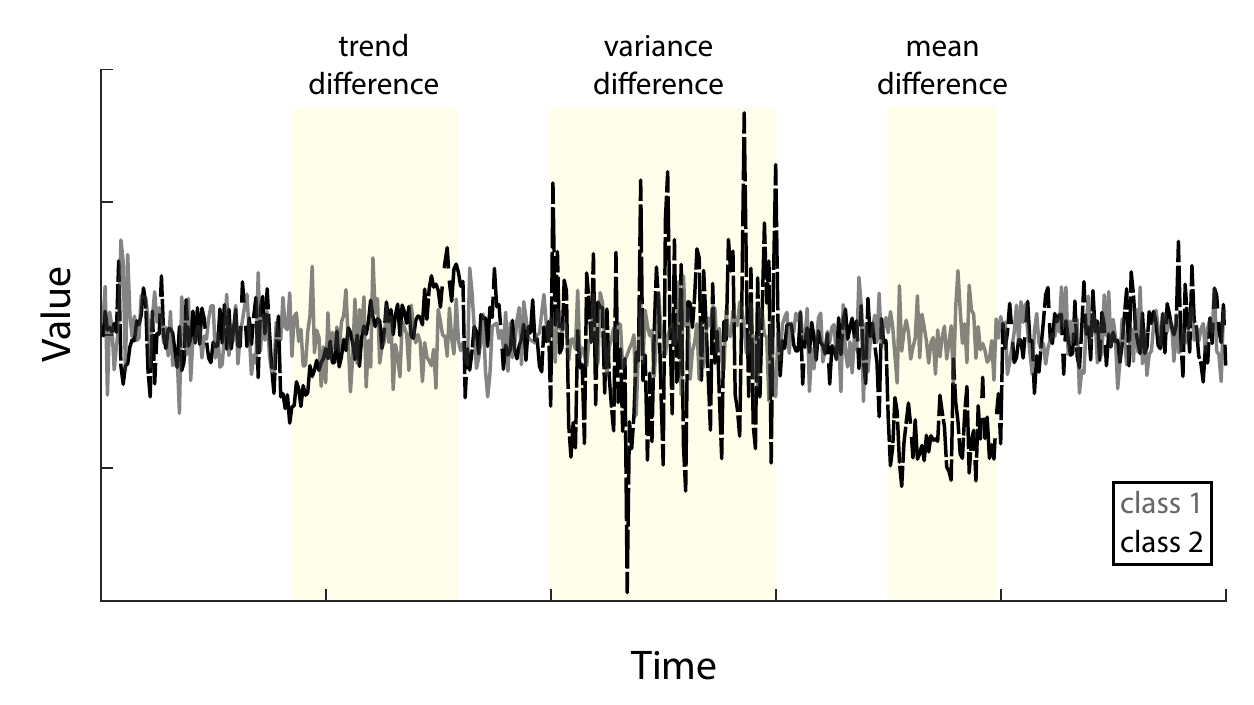}
  \caption[Interval features for time-series classification]{
  \textbf{Discriminating interval features}.
  Computing simple features in subsequences of a time series allows interpretable differences to be identified and used to distinguish labeled classes of time series.
  In this example, representative samples from two classes are shown, and periods of time in which they differ in their linear trend, variance, and mean, are highlighted.
  Quantifying these features and the specific time intervals at which they are most discriminative provides interpretable understanding of how, and when, the two data classes differ.
  }
  \label{fig:intervalFeatures}
\end{figure}

As depicted in Fig.~\ref{fig:intervalFeatures}, some time-series classification problems may involve class differences in time-series properties that are restricted to specific discriminative time intervals.
Interval classifiers seek to learn the location of discriminative subsequences and the features that separate different classes, which can be learned by computing simple features across many subsequences, and then building classifiers by searching over both features and time intervals \cite{Rodriguez01, Deng13}.
\index{time-series interval features}

Deng et al.~\cite{Deng13} used three simple features to capture the properties of time-series subsequences (for an interval $t_1 \leq x \leq t_2$) to aid interpretability and computational efficiency: the mean, $\frac{1}{t_2-t_1+1}\sum_{i=t_1}^{t_2} x_i$, standard deviation, cf. Eq.~\eqref{eq:acf}, and the slope (computed from a least squares regression line through the interval).
Differences in these properties are shown visually in Fig.~\ref{fig:intervalFeatures}.
In this way, each time-series subsequence is represented by the values of these three features, after which thresholds are learned on interval feature values using an entropy gain splitting criterion (and breaking ties by taking the split that maximizes the margin between classes).
After random sampling of time intervals and accumulating many decision trees, a resulting time-series forest classifier was used to classify new time series (as the majority vote from all individual decision trees).
To gain interpretable understanding, the contribution of each feature to the performance of the classifier at each time point was calculated (in terms of the entropy gain of that feature at a given time point), yielding an importance curve that indicates which of the three simple features contribute most to classification at each time point.
For example, for times at which spread differs between the classes, the standard deviation feature will have high temporal importance, but where there are differences in location between the two series, the mean will have high importance.
This information can be used to understand which time-series properties drive successful classification at each time point.
Recent work has used feature-feature covariance matrices to capture subsequence properties for classification \cite{ergezer2017time}.

\subsection{Shapelets}
\begin{figure}[h]
  \centering
    \includegraphics[width=0.7\textwidth]{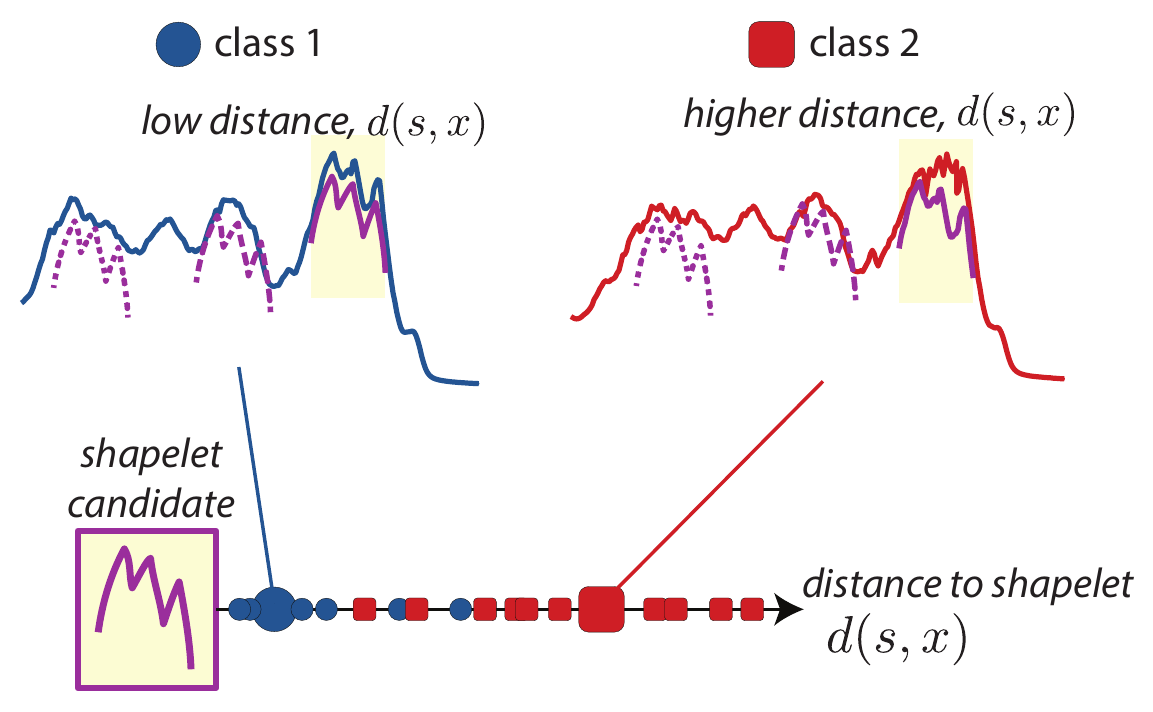}
  \caption[Shapelets]{
  \textbf{The distance between a time series, $x$, and a subsequence pattern, or \emph{shapelet}, $s$, or $d(s,x)$, can be used as the basis of classifying time series.}
The distance, $d(s,x)$, is defined as the minimum Euclidean distance across translation of the subsequence across the time series.
  In this example, the candidate shapelet is closer to time series of class 1 (picking up the shape of the peaks towards the end of the time series), but further from class 2, which has a different peak shape.
  An example time series from class 1 (blue circles) and class 2 (red squares) is plotted to illustrate [$d(s,x)$ for the two plotted examples are shown as a larger circle/square].
  The shapelet is an interpretable subsequence that can be used to quantify and understand subsequence patterns that differ between classes of time series.
  }
  \label{fig12:shapelets}
\end{figure}

Another representation for time series is in terms of the individual subsequences themselves.
In the context of time-series classification, subsequences that are highly predictive of class differences are known as \emph{shapelets} \cite{Ye09, mueen2011}, and provide interpretable information about the types of sequential patterns (or shapes) that are important to measure for a given problem.
\index{shapelets}
The problem of shapelet discovery can be framed around determining subsequences, $s$, that best distinguish different classes of time series by their distance to the shapelet, $d(s,x)$.
The distance between a time series, $x$, and a subsequence, $s$, of length $l$, can be defined as the minimum Euclidean distance across translation of the subsequence across the time series:
\begin{equation}
    d(s,x) = \min_k d(s,x_{k,...,k+l}),
\end{equation}
for a Euclidean distance function, $d$.
This distance, $d(s,x)$, can be thought of as the `feature' extracted from the time series.
The problem is illustrated in Fig.~\ref{fig12:shapelets}, where the depicted shapelet candidate captures the class difference in the shape of the peaks towards the end of the time series, which better matches time series of class 1, yielding a lower $d(s,x)$ for class 1.
In this way, shapelet-based classifiers can learn interpretable discriminative subsequences for time-series classification problems.
Shapelets:
(i) provide directly interpretable subsequence patterns that are discriminative of time-series classes,
(ii) allow efficient classification of new data using simple rules based on a small set of learned shapelets (avoiding the need to compare to a large training set of time series), and
(iii) ignore shapes in the time series that are not informative of class differences, thereby helping to improve generalization and robustness to noise.

In early work by Geurts \cite{Geurts01}, interpretable classification rules were learned based on patterns in subsequences after transforming time series piecewise constant representations.
Taking subsequences from time series in the dataset, Geurts noted the difficulty of determining the best subsequences from the vast space of all possible subsequences (across all possible subsequence lengths); instead taking subsequences from a single, randomly chosen instance from each class and used a piecewise constant representation to reduce the search space.
The problem was revisited by Ye and Keogh \cite{Ye09}, who introduced an algorithmic framework for feasibly searching across the massive space of all possible candidate time-series subsequences represented in a dataset.
They framed the classification problem in terms of the discovery of subsequences whose distance to a time series is informative of its class label, defining a \emph{shapelet} as the most informative such subsequence \cite{Ye09}.
Discriminability was quantified using information theory in one dimension, with a single threshold used to compute the \emph{optimal split point} as the split that maximizes the \emph{information gain} between classes, or using multiple splits in a decision tree framework for multiclass problems.
The framework has been applied widely and extended, for example to incorporate multiple shapelets (as logical combinations of individual shapelets, as \emph{and} and \emph{or} \cite{mueen2011}),
using genetic algorithms to improve subsequences \cite{sugimura2011},
and defining representative patterns as subsequences that appear frequently in a given class of time series and are discriminative between classes \cite{wang2016rpm}.
Lines et al.~\cite{lines12shapelet} implemented a shapelet transform that extracts a set of optimal shapelets, independent of any decision-tree based classification system, which could then be used to transform a given time series into a feature-based representation, where $k$ features are the set of distances to $k$ shapelets extracted from the dataset.
This shapelet distance feature-based representation of the dataset was then fed into a standard classification algorithm (like a random forest or support vector machine) \cite{lines12shapelet}.


\subsection{Pattern dictionaries}
\begin{figure}[h]
  \centering
    \includegraphics[width=.85\textwidth]{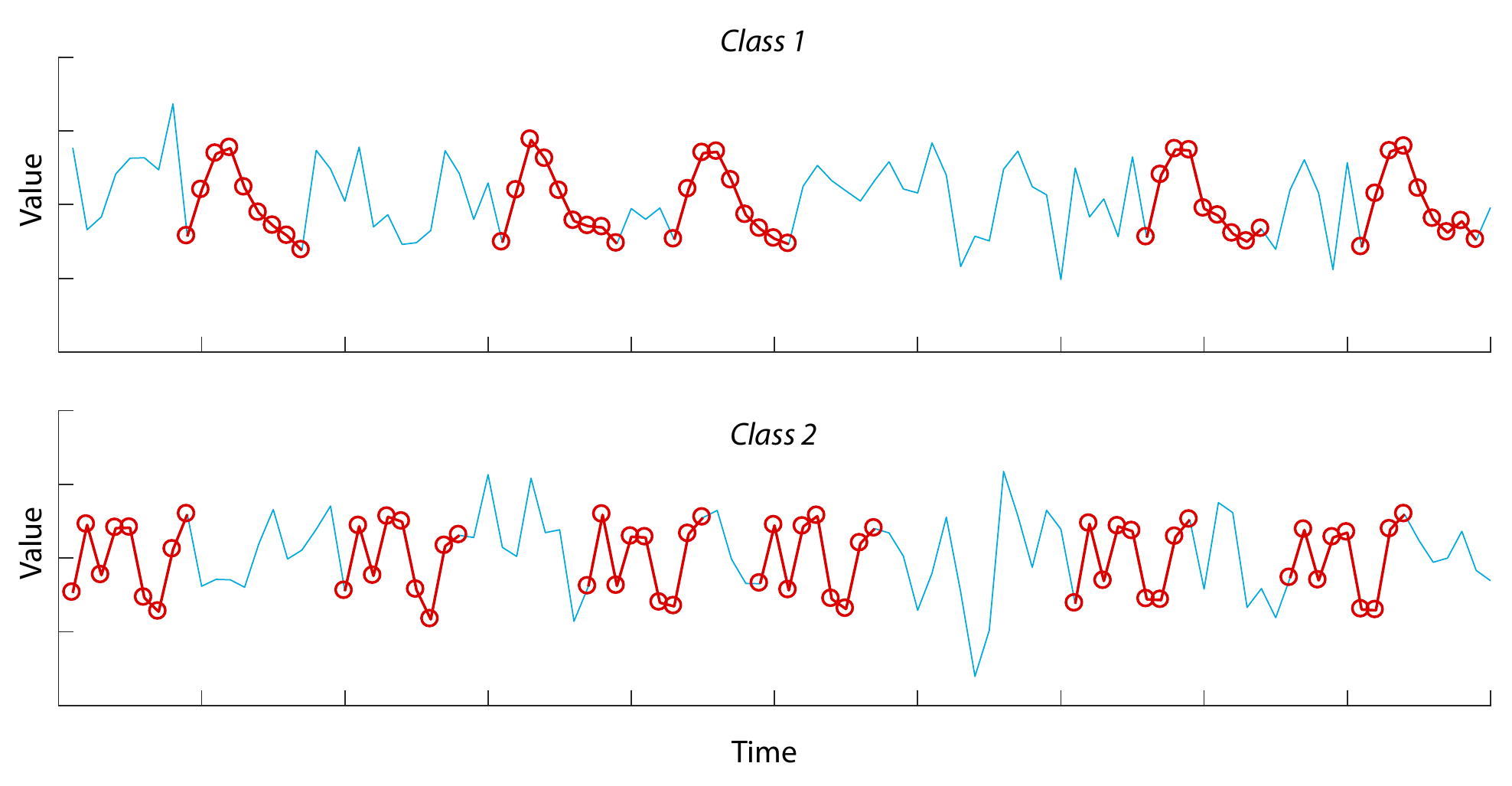}
  \caption[Time-series motifs for pattern dictionaries]{
  Capturing differences between two classes of time series by the frequency of shorter motifs.
  Characteristic motifs of each class are highlighted using circles in this idealized example; representing each class by the frequency of these motifs can be used as the basis for classification.
  }
  \label{fig:motifs}
\end{figure}

While shapelets can capture how well a subsequence matches to a time series (and are well-suited to short, pattern-based time series), they cannot capture how many times a given subsequence is represented across an extended time-series recording.
For example, consider two types of time series that differ in their frequency of short, characteristic subsequence patterns across the recording, depicted in Fig.~\ref{fig:motifs}.
In this case, class 1 has many occurrences of an increasing and then decreasing pattern (highlighted using circles), whereas class 2 features a characteristic oscillatory pattern.
Learning these discriminative patterns, and then characterizing each time series by the frequency of each pattern across the recording, provides useful information about the frequency of discriminative subsequences between classes of time series.
This representation is likely to be important in capturing stereotypical dynamic motifs, such as characteristic movement patterns in different strains of the nematode \emph{C. elegans} \cite{Bagnall:2015}.
In analogy to the bag-of-words representation of texts that judges two documents as similar that have similar relative frequencies of specific words, the time-series pattern dictionary approach judges pairs of time series as similar that contain similar frequencies of subsequence patterns.
\index{pattern dictionary}
It thus represents time series as a histogram that counts the number of matches to a given set of subsequence patterns across the full recording (ignoring their relative timing).


Variations on this basic approach have been used to tackle a range of problems.
The histogram-based bag of patterns approach of Lin et al.~\cite{lin2012rotation} represents time series using the frequency of shorter patterns, summed across a discretized SAX transformation of the time series \cite{Lin07}.
Sch\"after \cite{schafer2015boss} analyzed a symbolic representation of the time series in each window (based on a Symbolic Fourier Approximation), with an overall histogram across all sliding windows used to represent the time series as a whole.
Baydogan et al.~\cite{baydogan2013bag} computed mean, slope, and variance features in time-series subsequences (as Deng et al.~\cite{Deng13}), as well as start and end points for the subsequence, after which all classified subsequences were combined as histograms to form a codebook to represent and assign each time series.



\section{Combining time-series representations}
Thus far we have seen how the similarity of pairs of time series can be defined in terms of interpretable, extracted features that can provide different types of interpretable understanding of class differences (beyond simply computing differences in the time-series values across the full recording).
As well as global feature-based representations, we have also seen how meaningful differences can be quantified in terms of characteristic time-series subsequences.
Some time-series similarity measures do not fit into these two types of representations, such as the compression-based dissimilarity measure \cite{Keogh04}, which uses algorithms to compress each time series and defines time-series similarity based on the Kolmogorov complexity.
Other, hybrid approaches combine feature-based representations with conventional time-domain similarity measures.
For example, Batista et al.~\cite{batista2014cid} used a global feature of `complexity', $CE$, measured simply as the length of outstretched lines connecting successive points in a time series:
\begin{equation}
    CE = \sqrt{\sum_{i=1}^{N-1} (x_{i+1}-x_i)^2},
\end{equation}
for time-series, $x$, of length $N$, to re-weight Euclidean distances between pairs of time series.
In this way, pairs of time series are similar when they both: (i) have similar values through time, and (ii) have similar global `complexity'.
Kate \cite{kate2016using} used time-domain distances to form a feature-based representation of time series, as the set of (e.g., dynamic time warping, DTW) distances to a set of training time series.
\index{dynamic time warping (DTW)}

Comparing the large number of representations for time series, researchers have started to more comprehensively characterize which representations are better suited to which types of problems.
While it is conventionally the role of the researcher to select and interpret the data representation that provides the most interesting insights into the scientific question being asked of the data, promising recent research has combined multiple time-series representations into an ensemble, partially automating this subjective procedure.
For example, Bagnall et al.~\cite{Bagnall:2015} combined 34 classifiers based on four representations of time series:
(i) eleven time-domain representations (based on different elastic distance measures),
(ii) eight power spectrum classifiers,
(iii) eight autocorrelation-based classifiers, and
(iv) eight shapelet classifiers, in a simple ensemble named a Flat Collective Of Transformation-based Ensembles, or \emph{Flat-COTE}.
\index{Collective Of Transformation-based Ensembles (COTE)}
This was later extended as a Hierarchical Vote Collective Of Transformation-based Ensembles, or \emph{HIVE-COTE} \cite{lines2016hive}, which yields a single probabilistic prediction from five domains:
(i) time-domain,
(ii) time series forest based on simple interval features,
(iii) shapelets,
(iv) dictionary-based bag-of-SFA-symbols \cite{schafer2015boss},
and (v) random interval spectral features.
As we have seen, large collections of global time-series features can automate the selection of informative features for specific time-series problems \cite{Fulcher:2013ft, Fulcher:2014uo, Fulcher:2016ht, Christ:2016}.
Continuing in this direction, we may eventually be able to compare a large number of time-series representations automatically in order to understand which representations best suit a given problem.
The results of such a comparison would yield multiple interpretable perspectives on the dataset, rather than those obtained from a small number of manually-selected methods.

\section{Feature-based forecasting}

In the final section of this chapter, we describe the use of time-series features for tackling time-series forecasting, as depicted in Fig.~\ref{fig12:forecasting}.
\index{time-series forecasting}
Forecasting is typically tackled by fitting a statistical model to the data and then simulating it forward in time to make predictions.
A feature-based approach avoids training a model directly on sequences of time-series values, but instead uses features computed in reduced time intervals to make the prediction.
Many of the time-series features characterized above (in the context of time-series similarity measures for classification) could also be applied to forecasting problems.
For example, a feature-based approach to weather prediction by Paras et al.~\cite{paras2009} involved training neural networks on weighted averages, trend, and moments of the distribution in windows of time-series data.
Similarly, other work has used shapelets for forecasting \cite{xing2011extracting, mcgovern2011identifying}.
\index{shapelets}

Just as recent advances in time-series similarity metrics described above (e.g., for classification) have recognized that no single time-series representation can perform best on all problems \cite{wolpert1996, Wolpert97, bagnallreview}, in the context of forecasting, no single time-series representation or forecasting method can perform best on all types of time series.
\index{No-Free-Lunch theorem}
Recognizing this, hybrid approaches, such as \emph{rule-based forecasting}, first measure time-series features to characterize the data and then weight predictions of different time-series forecasting models accordingly \cite{collopy1992, Adya2001, Armstrong2001}.
This approach has been extended to learn relevant features from data using grammars, and then using feature selection and machine learning methods to generate predictions \cite{de2015grammar}.
Following this idea further, a detailed investigation was performed recently by Kang et al.~\cite{kang2017}, who questioned whether the types of time series studied in classic forecasting datasets represent the diversity (and therefore the true challenge) of forecasting new types of real-world data.
In this work, each time series is represented as a vector containing six features (spectral entropy, trend, seasonality, seasonal period, lag-1 autocorrelation, and the optimal Box-Cox transformation parameter), after which the full time-series dataset was projected to a two-dimensional principal components feature space in which time series with different properties occupy different parts of the space.
For example, some parts of the space contain time series with decreasing trend, where other parts contain time series with strong seasonality.
After embedding time-series data in a meaningful feature space, the performance of different forecasting algorithms was visualized in the space, providing an understanding of:
(i) which algorithms are best suited to which types of time series,
(ii) how the results may generalize to new datasets (in which time series may occupy different parts of the space), and
(iii) which types of time series are the most challenging to forecast, and therefore where future development of forecasting algorithms should be focused.
Perhaps most interestingly, Kang et al.~\cite{kang2017}, also introduce a method for correcting for potential bias in datasets of time series that may be overrepresented by time series of a certain type, i.e., by generating new time-series instances in sparse parts of the instance space.
The problem of visualizing and generating new time series with a given set of feature-based characteristics is ongoing \cite{kegelfeature}.

\section{Summary and outlook}
In this chapter, we have provided an overview of a vast literature of representations and analysis methods for time series.
We have encountered global distances between time-series values (including Euclidean and elastic distance measures like DTW), subsequences that provide more localized shape-based information, global features that capture higher order structure, and interval features that capture discriminative properties in time-series subsequences.
The most useful method for a given task is determined by the structure of the data (e.g., whether time series are of the same length, are phase-aligned, or whether class differences are global or restricted to specific intervals), and the context of the problem (e.g., whether accuracy or interpretability is more important, and what type of understanding would best address the domain question of interest).

We have seen how the large and interdisciplinary toolset for characterizing time-series properties has been adapted for problems ranging from classification, regression, clustering, forecasting, and time-series generation, as well as how more tailored approaches based on time-series subsequences have been developed.
A growing literature acknowledges that modern machine learning approaches can overcome some of the limitations of traditional data analysis, which is often plagued by subjective choices and small-scale comparison.
This includes the use of large, interdisciplinary databases of features that can be compared systematically based on their empirical performance to automate feature selection, for example \cite{Fulcher:2013ft, Fulcher:2014uo, Fulcher:2016ht, Christ:2016}, and the use of ensemble methods that try to understand the properties of a time series or time-series dataset that make it suitable for a particular representation or algorithm \cite{bagnall2012transformation, harvey2015automated, Bagnall:2015, kang2017}.
These approaches acknowledge that no algorithm can perform well on all datasets \cite{wolpert1996, Wolpert97}, and use modern statistical approaches to tailor our methods to our data.
While complex machine learning methods are sometimes criticized for being difficult to interpret, these examples show how feature-based statistical learning approaches can allow analysts to leverage the power and sophistication of diverse interdisciplinary methods to automatically glean interpretable understanding of their data.
The future of modern data analysis, including for problems involving time series, is likely to embrace such approaches that partially automate human learning and understanding of the complex dynamical patterns in the time series we measure from the world around us.


\end{document}